\documentclass{article}

\usepackage{microtype}
\usepackage{graphicx}
\usepackage{subcaption}
\usepackage{booktabs} 

\usepackage{hyperref}

\usepackage[preprint]{icml2026}

\usepackage{amsmath}
\usepackage{amssymb}
\usepackage{mathtools}
\usepackage{amsthm}

\usepackage[capitalize,noabbrev]{cleveref}

\theoremstyle{plain}

\theoremstyle{definition}

\theoremstyle{remark}

\usepackage[textsize=tiny]{todonotes}

\usepackage{xspace}
\usepackage{comment}
\usepackage[pdftex,dvipsnames]{xcolor}
\usepackage{tabularx}
\usepackage{booktabs}
\usepackage{arydshln} 
\usepackage{multirow}

\newcommand{\ourmethod}{\textsc{WAC}\xspace}

\newcommand\eg[0]{\textit{e.g.}}

\begin{document}

\twocolumn[
  \icmltitle{World-Model–Augmented Web Agents with Action Correction}



  \icmlsetsymbol{equal}{*}

  \begin{icmlauthorlist}
    \icmlauthor{Zhouzhou Shen}{equal,zju}
    \icmlauthor{Xueyu Hu}{equal,zju}
    \icmlauthor{Xiyun Li}{tenc}
    \icmlauthor{Tianqing Fang}{tenc}
    \icmlauthor{Juncheng Li}{zju}
    \icmlauthor{Shengyu Zhang}{zju}
  \end{icmlauthorlist}

  \icmlaffiliation{zju}{Zhejiang University}
  \icmlaffiliation{tenc}{Tencent AI Lab}

  \icmlcorrespondingauthor{Shengyu Zhang}{sy\_zhang@zju.edu.cn}

  \icmlkeywords{Machine Learning, GUI Agent}

  \vskip 0.3in
]



\printAffiliationsAndNotice{\icmlEqualContribution}

\begin{abstract}
Web agents based on large language models have demonstrated promising capability in automating web tasks.
However, current web agents struggle to reason out sensible actions due to the limitations of predicting environment changes, and might not possess comprehensive awareness of execution risks, prematurely performing risky actions that cause losses and lead to task failure.
To address these challenges, we propose \ourmethod, a web agent that integrates model collaboration, consequence simulation, and feedback-driven action refinement.
To overcome the cognitive isolation of individual models, we introduce a multi-agent collaboration process that enables an action model to consult a world model as a web-environment expert for strategic guidance; the action model then grounds these suggestions into executable actions, leveraging prior knowledge of environmental state transition dynamics to enhance candidate action proposal.
To achieve risk-aware resilient task execution, we introduce a two-stage deduction chain. A world model, specialized in environmental state transitions, simulates action outcomes, which a judge model then scrutinizes to trigger action corrective feedback when necessary.
Experiments show that \ourmethod achieves absolute gains of 1.8\% on VisualWebArena and 1.3\% on Online-Mind2Web.
\end{abstract}

\section{Introduction}

Large language models (LLMs), with powerful understanding and generation capabilities, have become the foundation of popular intelligent applications, including web agents \cite{pmlr-v70-shi17a,10.5555/3666122.3667845} for web task automation.
Some efforts, such as ReAct \cite{Yao2023}, attempt to construct a paradigm in which a web agent’s reasoning and execution are interleaved. This paradigm relies on a single model to perform reasoning, propose actions, and execute those actions, and has shown initial success on certain web-based tasks.

\begin{figure}[htbp]
  \centering
  \includegraphics[width=\columnwidth]{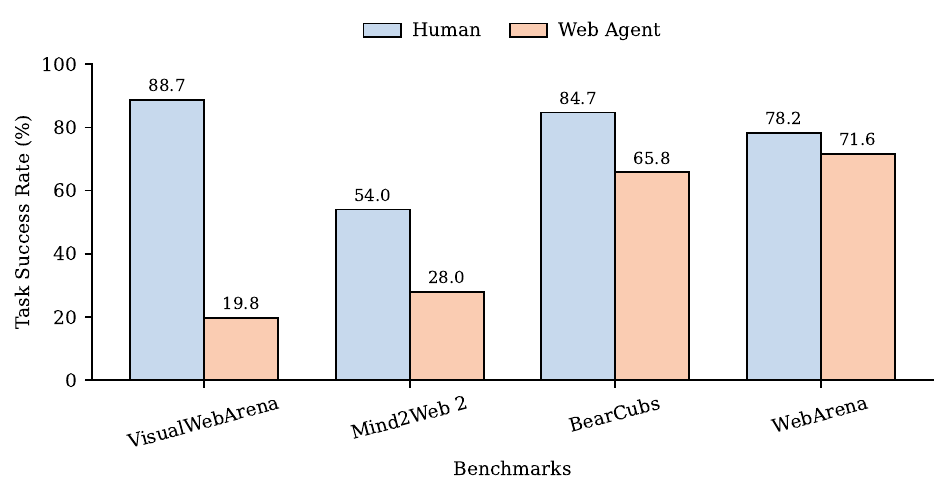}
  \caption{
    Human and web agent task success rates across representative web benchmarks. Despite recent progress, state-of-the-art web agents still substantially underperform humans.
  }
  \label{fig:web_agent_vs_human}
\end{figure}

Despite these advances, web agents still achieve substantially lower task success rates than humans on challenging benchmarks such as VisualWebArena \cite{koh2024visualwebarena}, as shown in \cref{fig:web_agent_vs_human}. 
Empirical evaluations reveal that web agents relying on single models frequently suffer from systematic reasoning limitations and may execute inadequate actions, leading to irreversible deviations in task trajectories.

Motivated by the limitations of existing web agents, we identify two key challenges that hinder their performance in complex web environments.
\textbf{(C1) Cognitive limitations of single-model decision making.}
Most existing web agents rely on a single model to simultaneously perceive the environment, reason about task progress, and generate executable actions.
However, the knowledge coverage and reasoning capacity of an individual model are inherently limited.
Agents may not adequately leverage complementary expertise or alternative perspectives during decision making, often leading to inadequate reasoning and suboptimal action proposals.
Such limitations can lead to error accumulation over time, which in turn significantly degrades overall success rates of web tasks.
\textbf{(C2) Lack of explicit pre-execution risk awareness and correction.}
Current web agents typically select and execute an action once it is proposed, without explicitly evaluating its potential consequences or associated risks.
Although some recent approaches incorporate outcome simulation to estimate action effects, they often lack mechanisms to reject or revise inadequate actions before execution.
When none of the candidate actions are satisfactory, the agent is still forced to commit to one, which can result in irreversible deviations in task trajectories.

To address these challenges, we introduce \ourmethod, short for \textbf{W}orld-model–augmented \textbf{A}ction \textbf{C}orrection, a web agent designed for collaboratively grounded action generation and risk-aware pre-execution action refinement.
To address C1, we equip \ourmethod with a multi-agent collaboration process that supports collaborative reasoning between two specialized models. Specifically, a world model endowed with extensive web environment state-transition knowledge acts as an environment expert, providing strategic guidance for single-step planning. This guidance is then integrated by an action model, which combines it with its own observations to deliberate and generate executable action proposals. Furthermore, to avoid unnecessary overhead from excessive collaboration, we introduce a collaboration-on-demand strategy. Collaboration is selectively triggered based on a confidence score, allowing the agent to invoke multi-agent reasoning only when needed and thereby improving both efficiency and action proposal quality.
To tackle C2, we design a world-model-centered deduction chain for pre-execution risk evaluation and closed-loop action correction. The same world model described in C1 is reused here in a different role: it simulates the potential outcomes of candidate actions to assess their effects and associated risks prior to execution. Based on these simulated outcomes, we introduce a closed-loop feedback mechanism that evaluates the feasibility of candidate actions and provides immediate corrective signals when the proposed actions are deemed inadequate. Upon receiving such feedback, the agent re-initiates action proposal and deduction, allowing decisions to be iteratively refined before execution.

We evaluate \ourmethod on two widely used benchmarks, VisualWebArena and Online-Mind2Web \cite{deng2023mind2web, xue2025an}, which feature diverse web environments and complex tasks. Across both benchmarks, \ourmethod consistently outperforms prior state-of-the-art decision-making workflows, achieving success rates of 24.5\% and 16.0\%, respectively. In addition, ablation studies and targeted analyses demonstrate that the collaboration process and the deduction chain effectively address the challenges corresponding to C1 and C2, respectively.

In summary, our contributions are three-fold:
\begin{itemize}
    \item We propose a novel web agent framework that integrates collaboratively grounded action generation and world-model-centered action deduction. By enabling on-demand model collaboration and a feedback-driven, closed-loop action refinement mechanism, our approach effectively mitigates the cognitive isolation of individual models (C1) and reduces premature commitment to risky or misaligned actions (C2).
    \item We conduct extensive experiments on VisualWebArena and Online-Mind2Web, demonstrating that our method consistently outperforms strong baselines. In particular, \ourmethod achieves absolute improvements of 1.8\% on VisualWebArena and 1.3\% on Online-Mind2Web, validating its effectiveness in enhancing action generation and refinement for web-based tasks.
    \item We provide comprehensive analyses, including ablation studies, cross-model generalization, and qualitative case studies, to better understand the behavior of our agent. These analyses reveal how collaborative action generation and feedback-driven refinement contribute to improved robustness.
\end{itemize}

\section{Related Work}

\subsection{LLM-Based Web Agent}

LLM–based web agents have seen rapid development recently.
Early works like RCI \cite{10.5555/3666122.3667845}, WebAgent \cite{ICLR2024_e91bf7df}, and WebGUM \cite{ICLR2024_7ef7d835} demonstrate the potential of language models to solve web-based tasks.
SeeAct \cite{zheng2023seeact}, TRISHUL \cite{Singh_2025_CVPR}, and WEPO \cite{Liu_Hao_Zhang_Hu_2025} further improve action grounding by leveraging multimodal visual–semantic alignment, enriched GUI representations, and distance-based preference optimization, respectively.
For reasoning and planning, prior works have explored step-level demonstration retrieval \cite{zhou2024trad}, workflow memory \cite{wang2024agentworkflowmemory}, dynamic task decomposition \cite{yang2025webdartdynamicdecompositionreplanning}, subtask-centric benchmarks \cite{srivastava2025warcbenchwebarchivebased}, and inference-time tree search \cite{koh2024tree} to enhance multi-step decision-making in web agents.
To measure progress in web agents, works such as WebArena \cite{ICLR2024_4410c071}, VisualWebArena \cite{koh2024visualwebarena} and Online-Mind2Web \cite{deng2023mind2web, xue2025an} establish realistic and comprehensive evaluation pipelines.
These advancements continuously enhance the understanding and decision-making capabilities of web agents.

\subsection{World Model in Web Agent}

World model originally aims to learn a predictive model of environment dynamics, enabling agents to understand how the environment works efficiently without relying solely on real-world interactions \cite{10.1145/122344.122377, https://doi.org/10.5281/zenodo.1207631}.
Therefore, world models are commonly used for sample-efficient learning \cite{wu2022daydreamer, jeong2025objectcentricworldmodellanguageguided, gao2025websynthesisworldmodelguidedmctsefficient}.
As for web agent, WMA web agent \cite{chae2025web} and WebDreamer \cite{Gu2025WebDreamer} introduce world models to simulate action execution for improved action selection.
WebEvolver \cite{fang-etal-2025-webevolver} introduces a co-evolutionary framework where a world model LLM jointly develops with the agent policy.
R-WoM \cite{mei2025rwomretrievalaugmentedworldmodel} augments LLM-based world model simulations with knowledge retrieval, addressing hallucination and outdated training knowledge.  
Our \ourmethod, in addition to using an LLM-based world model to simulate environment dynamics, further leverages its state-transition knowledge to inform candidate plan generation.

\section{\ourmethod}
\label{sec:method}

\begin{figure*}[t]
  \centering
  \includegraphics[width=0.65\textwidth]{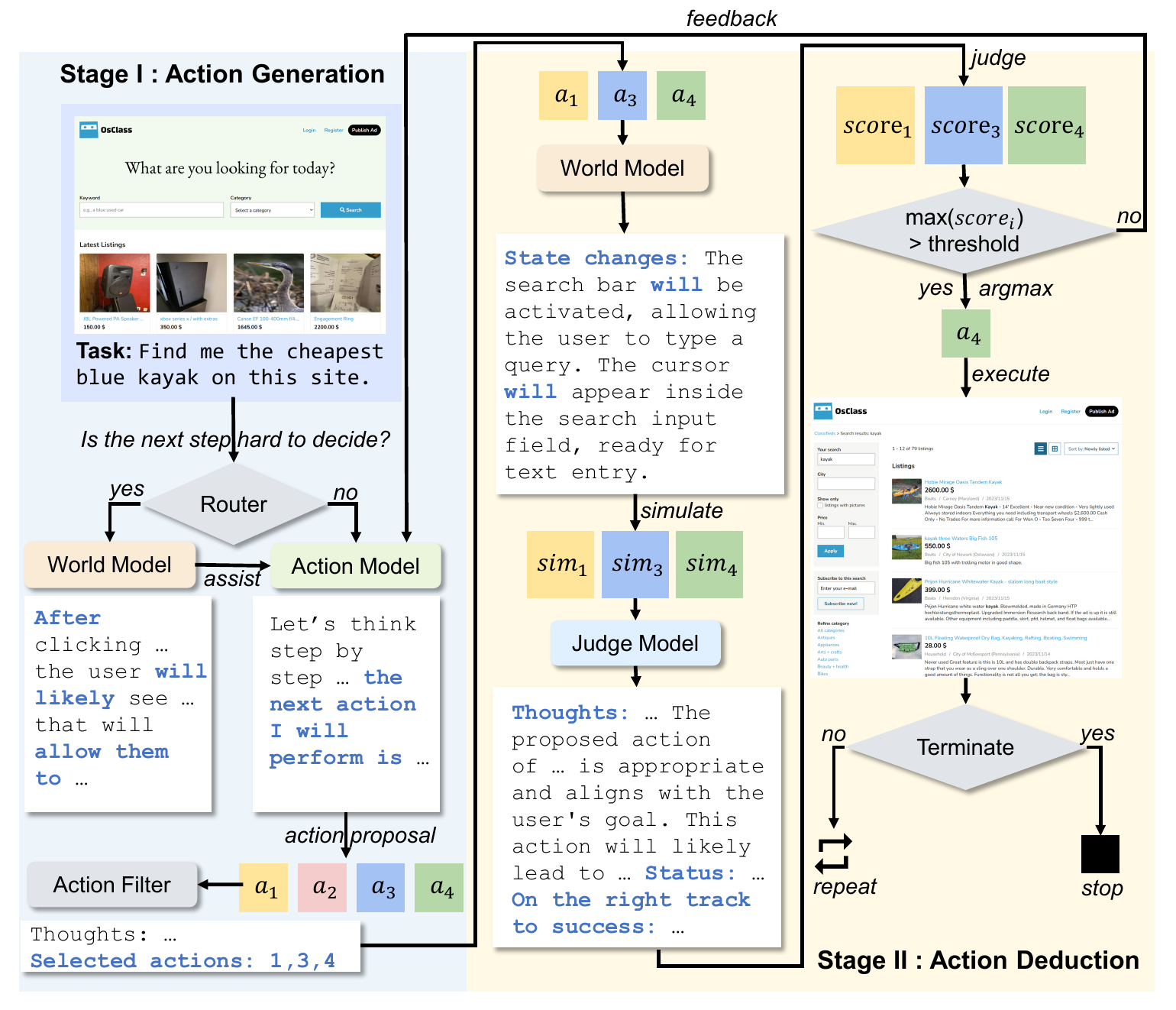}
  \caption{
    Overview of \ourmethod. Given the current observation and task, the agent first decides whether world-model assistance is needed for action generation. Candidate actions are then proposed and lightly filtered. Each candidate is simulated by a world model to predict potential state changes, which are evaluated by a judge model to assign confidence scores. If no action exceeds a predefined threshold, feedback derived from low-scoring simulations is used to refine action proposals in a closed loop. Once a high-confidence action is identified, it is executed in the environment, and the process repeats until termination.
  }
  \label{fig:overview}
\end{figure*}

\subsection{Problem Formulation}

Due to the partial observability of underlying webpage states, the stochasticity of web interactions, and the long-horizon sequential decision-making nature of web tasks, we model web agent task execution as a partially observable Markov decision process (POMDP), formally defined as $\left(\mathcal{S}, \mathcal{A}, \mathcal{O}, \mathcal{T}, \mathcal{Z}\right)$, where we focus on decision-making and execution under partial observability rather than reward optimization.
At time step $t$, the environment is in a latent state $s_t \in \mathcal{S}$, which is not directly accessible to the agent.
Instead, the agent receives an observation $o_t \in \mathcal{O}$ and selects an action $a_t \in \mathcal{A}$.
The environment then transitions to a new latent state $s_{t+1} \sim \mathcal{T}(\cdot \mid s_t, a_t)$, from which the next observation is generated as $o_{t+1} \sim \mathcal{Z}(\cdot \mid s_{t+1})$.
An episode terminates either when the agent explicitly decides to stop, or is passively terminated due to execution errors, excessive repeated actions, or reaching a maximum step limit.

\subsection{Framework Overview}

\begin{algorithm}[tb]
\caption{Workflow of \ourmethod}
\label{alg:ours}
\begin{algorithmic}[1]
\STATE {\bfseries Input:} Task $\tau$, Initial observation $o_0$
\STATE {\bfseries Output:} Executed action sequence $\{a_0, a_1, \dots, a_T\}$
\STATE Initialize execution history $\mathcal{H}_0 \leftarrow \emptyset$, $t \leftarrow 0$
\WHILE{termination\_check() = False}
    \STATE \textbf{// Stage I: Action Generation}
    \STATE $g_t \leftarrow \textsc{Router}(o_t, \tau)$
    \IF{$g_t = 1$}
        \STATE $s_t \leftarrow \mathcal{W}_{\text{guide}}(o_t, \tau, \mathcal{H}_t)$
    \ELSE
        \STATE $s_t \leftarrow \varnothing$
    \ENDIF
    \STATE $\tilde{\mathcal{A}}_t \leftarrow \textsc{ActionModel}(o_t, \tau, \mathcal{H}_t, s_t)$
    \STATE $\mathcal{A}_t \leftarrow \textsc{ActionFilter}(\tilde{\mathcal{A}}_t)$
    
    \STATE \textbf{// Stage II: Action Deduction}
    \STATE Initialize refinement counter $k \leftarrow 0$
    \REPEAT
        \FORALL{$a_t^{(i)} \in \mathcal{A}_t$}
            \STATE $\widehat{o}_{t+1}^{(i)} \leftarrow \mathcal{W}_{\text{sim}}(o_t, a_t^{(i)})$
            \STATE $(c^{(i)}, r^{(i)}) \leftarrow \textsc{Judge}(\widehat{o}_{t+1}^{(i)}, \tau, \mathcal{H}_t)$
        \ENDFOR
        \IF{$\max_i c^{(i)} \ge \theta$}
            \STATE $a_t^\star \leftarrow \arg\max_i c^{(i)}$
            \STATE \textbf{break}
        \ENDIF
        \STATE $\mathcal{G}_t \leftarrow \{(a_t^{(i)}, r^{(i)}) \mid c^{(i)} < \theta\}$
        \STATE $\mathcal{A}_t \leftarrow \textsc{Refine}(\mathcal{A}_t, \mathcal{G}_t)$
        \STATE $k \leftarrow k + 1$
    \UNTIL{$k = K_{\max}$}
    \IF{$\max_i c^{(i)} < \theta$}
        \STATE $a_t^\star \leftarrow \arg\max_i c^{(i)}$
    \ENDIF
    
    \STATE $o_{t+1} \leftarrow \textsc{Execute}(a_t^\star)$
    \STATE $\mathcal{H}_{t+1} \leftarrow \mathcal{H}_t \cup \{(o_t, a_t^\star)\}$
    \STATE $t \leftarrow t + 1$
\ENDWHILE
\end{algorithmic}
\end{algorithm}

We present \ourmethod, a multi-agent web agent framework designed to support collaborative action generation and risk-aware pre-execution action refinement.
Unlike prior ReAct-style \cite{Yao2023} agents that rely on a single model to both reason and act, \ourmethod decomposes decision-making into complementary stages with specialized model roles.
As illustrated in \cref{fig:overview}, \ourmethod consists of two sequential stages: \textbf{Action Generation} and \textbf{Action Deduction}.

In \textbf{Stage~I (Action Generation)}, the agent produces a set of candidate actions grounded in the current web observation. 
To mitigate the limited reasoning capacity of a single model, the action model may collaborate with the world model, which acts as an environment expert to provide strategic guidance for action proposal. 
This collaboration is triggered on demand when the agent lacks sufficient confidence to generate reliable actions based solely on the action model.

In \textbf{Stage~II (Action Deduction)}, the agent evaluates the candidate actions prior to execution and performs feedback-driven refinement based on simulated execution outcomes.
The world model simulates the potential outcomes of proposed actions, which are then assessed by a judge model to provide feedback for pre-execution action refinement.
When all candidate actions are deemed inadequate or risky, corrective feedback is returned to the action generation process, enabling iterative refinement before execution.

Algorithm~\ref{alg:ours} summarizes the overall workflow of \ourmethod, illustrating how collaborative action generation and feedback-driven action refinement are integrated into a unified decision-making loop across execution steps.

\subsection{Action Generation}

Existing web agents (\eg, ReAct) typically rely on a single model to both reason about the task and to propose executable actions. 
In this case, the model's internal knowledge and reasoning capacity may be insufficient to capture environment dynamics (\eg, navigation rules, state dependencies), producing inadequate action proposals.
Recent multi-agent works endow multiple specialized agents with complementary knowledge and coordination mechanisms, raising a promising solution to the above issue.
Inspired by this, we design the Action Generation stage to produce a compact, environment-aware candidate action set \(\mathcal{A}_t\) through three steps: (1) collaboration-on-demand with a world model that supplies high-level guidance, (2) initial action proposal, and (3) lightweight action filtering.

At each step $t$ of web task execution, a LLM-based gating module determines---based on the current environment observation $o_t$ and the user-specified task $\tau$---whether the next action decision should be augmented with assistance from a world model endowed with extensive knowledge of web-environment state transitions. 
Formally, we denote the gating decision at step $t$ as
\begin{equation}
g_t = \textsc{Router}(o_t, \tau) \in \{0,1\},
\end{equation}
where $g_t=1$ indicates that world-model guidance is invoked.
If so, the world model---conditioned on $o_t$, $\tau$, and the interleaved execution trajectory $\mathcal{H}_t$ of past actions and webpage screenshots---performs prospective reasoning and outputs natural-language, actionable suggestions $s_t$ for the next step.
Subsequently, an action model conditions on the task $\tau$, the current observation $o_t$, the execution history $\mathcal{H}_t$, and the suggestion $s_t$ (when available) to produce its own reasoning process and a structured action decision.
Under probabilistic decoding, the action model is sampled multiple times to produce a set of candidate actions $\tilde{\mathcal{A}}_t = \{ a_t^{(1)}, \dots, a_t^{(N)} \}$. 
Following prior work \cite{Gu2025WebDreamer, chae2025web}, we use a LLM to perform lightweight filtering on $\tilde{\mathcal{A}}_t$ to remove clearly unreasonable candidate actions, yielding $\mathcal{A}_t$.

Leveraging the LLM-based world model to guide action proposal narrows the decision space to affordance-aligned moves and injects transition knowledge directly into the reasoning process. This reduces inadequate proposal and improves goal-directedness  in ambiguous execution steps. 

\subsection{Action Deduction}

To assess the risks of candidate actions, recent works \cite{Gu2025WebDreamer, chae2025web}, instead of directly scoring them, leverage a world model to simulate their potential outcomes and select actions based on these simulations.
However, when none of the candidate actions are appropriate, the agent is still forced to commit to one for execution, which can lead to irreversible consequences or deviations in the task trajectory that are difficult for existing agents to recover from.
Given that the task success rates of current web agents remain substantially lower than those of humans, such situations occur frequently in practice.
To overcome this issue, we propose a feedback-driven action refinement loop based on simulated execution outcomes.

At time step $t$, the Action Generation stage produces a set of candidate actions $\mathcal{A}_t = \{ a_t^{(1)}, \dots, a_t^{(M)} \}$ under the current observation $o_t$.
To evaluate the potential risks and effectiveness of each candidate prior to execution, we employ a world model $\mathcal{W}$ to simulate execution outcomes.
For each action $a_t^{(i)} \in \mathcal{A}_t$, the world model predicts a hypothetical next observation:
\begin{equation}
\widehat{o}_{t+1}^{(i)} = \mathcal{W}(o_t, a_t^{(i)}),
\end{equation}
where $\widehat{o}_{t+1}^{(i)}$ is a natural-language description of the simulated post-execution state.
Following WebDreamer \cite{Gu2025WebDreamer}, we set the world model simulation horizon to 1.

The simulated outcomes are then assessed by a judge model, which inspects each $\widehat{o}_{t+1}^{(i)}$ to produce:
(i) a discrete confidence score $c^{(i)} \in \{0, 0.5, 1\}$, indicating whether the action is judged to fail, partially complete, or successfully complete the task; and
(ii) a textual rationale $r^{(i)}$ explaining the evaluation.
For stability and consistency, the confidence score is obtained by quantifying the judge model’s answers to a set of predefined discrete questions, rather than being directly generated.
We denote the judge model as a function producing confidence--rationale pairs $(c^{(i)}, r^{(i)})$ for each simulated outcome.

If $\max_i c^{(i)} \ge \theta$, where $\theta$ is a predefined confidence threshold, the agent executes the highest-scoring action $a_t^{(i^\star)}$ with $i^\star = \arg\max_i c^{(i)}$.
Otherwise, when all candidate actions are assessed as low-confidence, we activate a feedback-driven action refinement loop.

Specifically, we pair inadequate candidate actions with their corresponding rationales to form corrective guidance, which is fed back to the action model as additional context.
Formally, We aggregate low-confidence actions and their rationales into
\begin{equation}
\mathcal{G}_t = \{(a_t^{(i)}, r^{(i)}) \mid c^{(i)} < \theta\},
\end{equation}
which serves as corrective guidance for action regeneration.
The action model then regenerates a refined candidate set under these constraints.
To retain relative strong candidates from previous iterations while incorporating newly generated actions, we apply an operator that merges historical and newly proposed actions and selects a compact subset for the next iteration.
Formally, let $\mathcal{A}_t^{\text{new}}$ denote the candidate actions newly generated by the action model, and let $\mathcal{A}_t^{\text{prev}}$ denote the retained action set from previous iterations.
We define the operator as:
\begin{equation}
\operatorname{Top}_N\!\left(
\mathcal{A}_t^{\text{prev}} \right) \cup \mathcal{A}_t^{\text{new}},
\end{equation}
where $\operatorname{Top}_N(\cdot)$ selects the $N$ actions with the highest confidence scores assigned by the judge model.
This design prevents the refinement loop from completely discarding previous actions, while progressively improving the candidate set through corrective feedback.
This process iterates in a closed loop of \emph{proposal $\rightarrow$ simulation $\rightarrow$ evaluation}, and terminates either when a high-confidence action emerges or when a maximum number of refinement iterations is reached.
We cap the refinement loop at $K_{\max}$ iterations for efficiency.
If the loop terminates due to iteration limits, the action with the highest confidence score among all previously evaluated candidates is selected for execution.

By preventing premature execution of intent-misaligned or implausible actions and exposing corrective signals before acting, this feedback mechanism improves robustness and reduces error accumulation in complex web tasks.

\section{Experiments}

\subsection{Experimental Settings}

We evaluate \ourmethod on two widely used web agent benchmarks: VisualWebArena \cite{koh2024visualwebarena} and Online-Mind2Web \cite{deng2023mind2web, xue2025an}.

\textbf{VisualWebArena} is a realistic benchmark for multimodal web agents, featuring tasks that require interpreting screenshots in addition to DOM and textual information. The benchmark is evaluated through execution on functional websites in sandbox environments. It consists of three sub-websites: Classifieds, Shopping, and Reddit. Performance is measured by success rate, defined as the proportion of tasks successfully completed according to the dataset’s evaluation rules. To reduce evaluation cost and maintain consistency with prior work, we evaluate on 233 human-verified tasks selected from the dataset.

\textbf{Online-Mind2Web} is the online variant of Mind2Web, featuring 300 tasks spanning 136 popular real-world websites and designed to evaluate agents directly on the live web. Mind2Web provides the original large-scale task set and schema. This benchmark adopts an LLM-as-a-judge evaluation pipeline to determine task success, achieving an 85\% agreement rate with human judgments. Performance is measured by success rate, defined as the proportion of tasks judged as successful.

\begin{figure}[htbp]
  \centering
  \includegraphics[width=0.6\columnwidth]{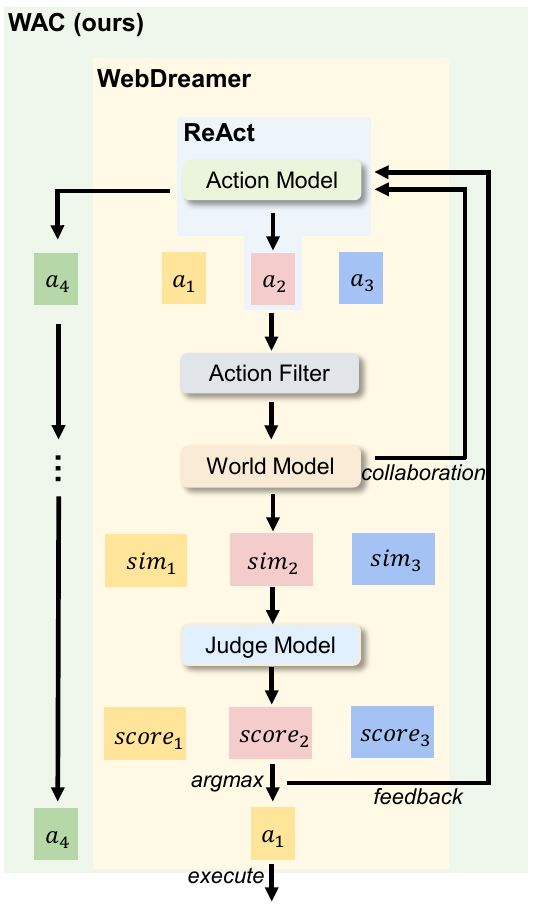}
  \caption{
    Comparison of action generation and pre-execution decision processes used by ReAct, WebDreamer, and our method (\ourmethod). While ReAct directly executes a single proposed action and WebDreamer selects among simulated candidates, \ourmethod enables collaborative action generation and feedback-driven action refinement prior to execution, leading to more robust action choices.
  }
  \label{fig:method_comparison}
\end{figure}

To ensure a fair comparison and demonstrate the effectiveness of \ourmethod, we adopt ReAct \cite{Yao2023} and WebDreamer \cite{Gu2025WebDreamer} as baseline methods and reimplement them on the selected benchmarks.
We include a visual comparison of decision-making behaviors to clarify how these methods differ during action selection, which is not intended to introduce new components beyond those described in \cref{sec:method}.
\cref{fig:method_comparison} provides the high-level comparison of the action generation and pre-execution decision processes used by these methods, highlighting the key behavioral differences that motivate our experimental evaluation.
\textbf{ReAct} is a strong reasoning-and-acting baseline that interleaves natural language reasoning with action execution, enabling agents to make decisions step by step based on observations.
\textbf{WebDreamer} extends this paradigm by incorporating a world model to simulate the outcomes of candidate actions before execution, allowing the agent to perform lookahead reasoning in web environments.
In contrast, \ourmethod further enables collaborative action generation and feedback-driven refinement, allowing suboptimal actions to be corrected before execution.

We implement all agents using the toolkit provided by VisualWebArena to enable interaction with web environments.
To ensure fair comparison with prior work, we faithfully reproduce the ReAct and WebDreamer baselines following their original implementations.
All agents are instantiated with the same vision–language model, Qwen-VL-Plus \cite{Qwen3-VL,Qwen2.5-VL,Qwen2-VL,Qwen-VL}, and the decoding temperature is fixed to 1 across all experiments.
For consistency with prior settings, we set the simulation horizon of the world model to 1, which has been empirically shown to be both effective and efficient.

\subsection{Main Results}

\begin{table*}[t]
    \centering
    \caption{Results on VisualWebArena and Online-Mind2Web.}
    \label{tab:main_results}
    \small
    \setlength{\tabcolsep}{10pt}
    \renewcommand{\arraystretch}{1.15}
    \begin{tabular}{clcc}
        \toprule
        \textbf{Model} & \textbf{Method} & \textbf{VisualWebArena} & \textbf{Online-Mind2Web} \\
        \midrule
        \multirow{3}{*}{Qwen3-VL-Plus}
            & ReAct & 22.7 & 12.7 \\
            & WebDreamer & 22.7 & 14.7 \\
            & \ourmethod (ours) & \textbf{24.5} & \textbf{16.0} \\
        \bottomrule
    \end{tabular}
\end{table*}

\cref{tab:main_results} reports task success rates on VisualWebArena and Online-Mind2Web using the same backbone model (Qwen3-VL-plus) across methods. 
Our method, \ourmethod, outperforms both ReAct and WebDreamer on both benchmarks. 
On VisualWebArena, \ourmethod achieves 24.5\% (vs.\ 22.7\% for both ReAct and WebDreamer), corresponding to a 7.9\% relative improvement over the two baselines.
On Online-Mind2Web, \ourmethod reaches 16.0\%, improving upon WebDreamer by 8.8\% in relative terms.

These results indicate that integrating collaborative action generation, which incorporates complementary suggestions from multiple models, together with feedback-driven pre-execution action refinement based on simulated outcomes, consistently improves decision quality and execution robustness across diverse web scenarios.
The consistent improvements on both benchmarks demonstrate the effectiveness of our approach for web-based decision making, where inadequate actions can lead to hard-to-recover states and thus call for careful action generation together with explicit pre-execution action validation and refinement.

\subsection{Ablation Studies}

\begin{table}[H]
    \centering
    \caption{Ablation study on VisualWebArena Classifieds.}
    \label{tab:ablation}
    \small
    \setlength{\tabcolsep}{6pt}
    \renewcommand{\arraystretch}{1.15}
    \begin{tabularx}{\columnwidth}{c>{\raggedright\arraybackslash}Xc}
        \toprule
        \textbf{Model} & \textbf{Method} & \textbf{SR$\uparrow$} \\
        \midrule
        \multirow{4}{*}{Qwen3-VL-Plus}
            & ReAct & 30.4 \\
            & WebDreamer & 32.1 \\
        \addlinespace[2pt]
        \cdashline{2-3}
        \addlinespace[4pt]
            & + Feedback-Driven Action Refinement & 33.9 \\
        \addlinespace[2pt]
        \cdashline{2-3}
        \addlinespace[4pt]
            & + Collaborative Action Generation & \textbf{35.7} \\
        \bottomrule
    \end{tabularx}
\end{table}

To evaluate the impact of Feedback-Driven Action Refinement and Collaborative Action Generation on agent performance, we conduct comprehensive experiments on the human-verified Classifieds subset of VisualWebArena, as shown in \cref{tab:ablation}.
We report the task success rates of ReAct, WebDreamer, and WebDreamer augmented with our proposed modules on 56 tasks from the human-verified Classifieds subset.
We observe a consistent improvement in task success rates as our proposed modules are progressively incorporated into the agent.
During task execution, Feedback-Driven Action Refinement helps correct implausible or intent-misaligned actions before execution, while Collaborative Action Generation enhances action proposal by incorporating environment-aware guidance from the world model.
Together, these components yield an absolute improvement of 3.6\% in task success rate, demonstrating the effectiveness of our methods.

\subsection{Further Analysis}

\paragraph{Cross-Model Generalization}

\begin{table}[H]
    \centering
    \caption{Cross-model generalization results on VisualWebArena Classifieds.}
    \label{tab:cross_model_classifieds}
    \small
    \setlength{\tabcolsep}{6pt}
    \renewcommand{\arraystretch}{1.15}
    \begin{tabularx}{\columnwidth}{c>{\raggedright\arraybackslash}Xc}
        \toprule
        \textbf{Model} & \textbf{Method} & \textbf{SR$\uparrow$} \\
        \midrule
        \multirow{3}{*}{GLM-4.5v}
            & ReAct & 21.4 \\
            & WebDreamer & 33.9 \\
            & \ourmethod (ours) & \textbf{37.5} \\
        \bottomrule
    \end{tabularx}
\end{table}

To assess whether our method generalizes beyond the foundation model used in previous experiments, we conduct a cross-model evaluation on the Classifieds subset of VisualWebArena by replacing the underlying foundation model with GLM-4.5v \cite{vteam2025glm45vglm41vthinkingversatilemultimodal}.
All methods are evaluated under identical environment and task settings to ensure a fair comparison.
As shown in \cref{tab:cross_model_classifieds}, \ourmethod consistently outperforms the baselines under the GLM-4.5v backbone, achieving a task success rate of 37.5, compared to 33.9 for WebDreamer and 21.4 for ReAct.
This performance gap suggests that the gains of \ourmethod are not tightly coupled to a specific foundation model, but instead arise from the proposed collaborative action generation and feedback-driven action refinement mechanisms, highlighting its potential applicability in practical web agent systems where model availability and characteristics may vary.

\begin{table}[H]
    \centering
    \caption{Cross-model generalization results on VisualWebArena Reddit.}
    \label{tab:cross_model_reddit}
    \small
    \setlength{\tabcolsep}{6pt}
    \renewcommand{\arraystretch}{1.15}
    \begin{tabularx}{\columnwidth}{c>{\raggedright\arraybackslash}Xc}
        \toprule
        \textbf{Model} & \textbf{Method} & \textbf{SR$\uparrow$} \\
        \midrule
        \multirow{3}{*}{GLM-4.6v}
            & ReAct & 7.9 \\
            & WebDreamer & 12.7 \\
            & \ourmethod (ours) & \textbf{14.3} \\
        \bottomrule
    \end{tabularx}
\end{table}

We further evaluate cross-model generalization on the VisualWebArena Reddit subset using the recently released GLM-4.6v backbone.
Reddit tasks are widely regarded as more challenging due to dense page layouts, high information volume, and the need for multi-step interactions, which lead to lower overall success rates for existing web agents.
As shown in \cref{tab:cross_model_reddit}, \ourmethod achieves a task success rate of 14.3, outperforming WebDreamer (12.7) and ReAct (7.9) by a clear margin.
Together, these results indicate that our method not only generalizes across different foundation models, but also remains effective under more challenging web scenarios, where careful action proposal and pre-execution action validation are especially important.

\paragraph{Can Feedback Replace Action Sampling?}

\begin{table}[H]
    \centering
    \caption{Performance of the single-proposal variant of \ourmethod on VisualWebArena Classifieds.}
    \label{tab:single_proposal}
    \small
    \setlength{\tabcolsep}{6pt}
    \renewcommand{\arraystretch}{1.15}
    \begin{tabularx}{\columnwidth}{c>{\raggedright\arraybackslash}Xc}
        \toprule
        \textbf{Model} & \textbf{Method} & \textbf{SR$\uparrow$} \\
        \midrule
        \multirow{2}{*}{Qwen3-VL-Plus}
            & \ourmethod & \textbf{35.7} \\
            & \ourmethod (single-proposal) & \textbf{35.7} \\
        \bottomrule
    \end{tabularx}
\end{table}

Existing web agents typically rely on repeated prompting and stochastic decoding to obtain a diverse set of candidate actions.
While effective, this strategy incurs substantial computational overhead and treats action diversity as a byproduct of sampling rather than structured refinement.
In contrast, \ourmethod introduces a feedback-driven refinement loop that iteratively corrects inadequate actions based on simulated execution outcomes.
This design raises a natural question: \textit{Can the feedback mechanism compensate for reduced action sampling?}
We preliminarily explore this question.
Specifically, we initialize $\mathcal{A}_t$ with a single action sampled from the action model, and rely solely on the feedback-driven refinement loop to iteratively improve the proposal.
As shown in \cref{tab:single_proposal}, despite using only a single initial action proposal, the variant maintains competitive performance, indicating that feedback-driven refinement may partially substitute for explicit action sampling.
This result highlights the potential of feedback-driven refinement as a more structured and efficient alternative to extensive action sampling, especially in latency- or cost-sensitive settings.

\paragraph{Effective Case Illustration}

\begin{figure*}[t]
  \centering
  \includegraphics[width=\textwidth]{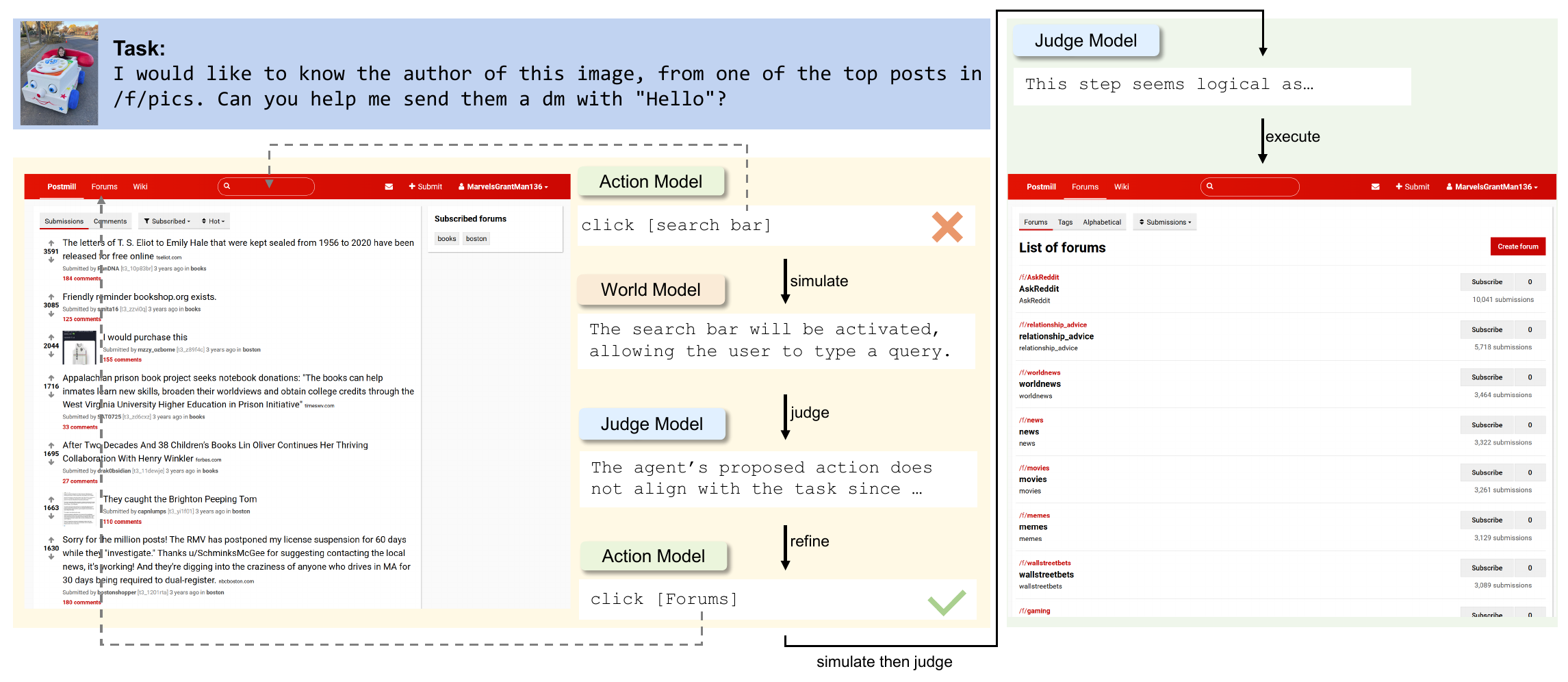}
  \caption{
    An illustrative case where feedback-driven action refinement corrects an initially proposed action at the first execution step. Although only a single candidate action is initially generated, simulated outcome evaluation identifies it as risky, triggering refinement and leading to a revised action that places the agent on a successful trajectory.
  }
  \label{fig:case_study}
\end{figure*}

We present a case study illustrating how \ourmethod corrects a inappropriate action at the very first step of task execution through feedback-driven action refinement.
For clarity, we omit the collaborative action generation process in the figure and focus on the refinement dynamics; in this case, the world model does not provide additional guidance at the initial step, and the action is proposed directly by the action model.
Given the task and the initial observation, the action model produces only a single candidate action at step $t=1$.
Instead of executing this action directly, \ourmethod leverages the environment-aware world model to simulate its potential outcome.
The simulated post-execution state reveals that this action would lead the agent onto an undesirable trajectory that is misaligned with the task objective.
Based on this simulated outcome, the judge model provides a rationale indicating the underlying issue.
Conditioned on this feedback, the action model revises its proposal and generates an alternative action that better respects the task requirements and page structure.
Executing this refined action places the agent on a correct execution trajectory, from which the task can be completed successfully.
This case highlights that even when action diversity is minimal and no alternative candidates are initially available, the proposed feedback-driven refinement loop can still intervene effectively.
By identifying and correcting a flawed action before execution, \ourmethod prevents early-stage errors that could otherwise propagate and become difficult to recover from in later steps.

\section{Limitations}

Despite the promising results achieved by \ourmethod, it still has several limitations: 
(1) Our approach operates in a tuning-free manner and therefore inherits the limitations of the underlying foundation models. 
In particular, the agent’s performance is influenced by the base model’s knowledge coverage, reasoning capability, and perceptual accuracy, which may constrain its effectiveness in certain scenarios.
(2) Although \ourmethod incorporates an environment-aware world model to simulate potential outcomes, such simulations are inherently approximate. 
Discrepancies between predicted and actual web environment dynamics may lead to imperfect feedback and suboptimal refinements in some cases.
In the future, we will try to enhance \ourmethod by incorporating lightweight adaptation or specialization techniques to better align the agent with target web environments while preserving its tuning-free advantages. 
We also plan to improve the fidelity of outcome simulation by leveraging stronger or dynamically updated world models to provide more accurate feedback signals.

\section{Conclusion}

In this work, we introduce \ourmethod, a web agent that integrates collaborative action generation and feedback-driven pre-execution action refinement.
By incorporating an environment-aware world model for action generation assistance and outcome simulation, together with a feedback mechanism for pre-execution action refinement, \ourmethod enables more informed action proposals and robust decision making in complex web environments.
This design mitigates structural limitations of single-model action reasoning and provides a closed-loop corrective mechanism that improves both action quality and execution robustness.
Extensive experiments on VisualWebArena and Online-Mind2Web show that \ourmethod consistently outperforms strong baselines, achieving task success rates of 24.5\% and 16.0\%, respectively.
We will open-source our code upon publication to facilitate reproducibility and future research in this domain.

\section*{Impact Statement}

This paper proposes a web agent framework that improves decision robustness through collaborative action generation and feedback-driven action refinement.
The primary goal of this work is to advance research on reliable and generalizable web agents capable of operating in complex, dynamic online environments.
From a positive perspective, such agents may assist users in information retrieval, task automation, and accessibility, potentially reducing the cognitive and operational burden of interacting with large-scale web systems.
By emphasizing pre-execution verification and iterative correction, our approach also promotes safer and more cautious agent behavior compared to directly executing unverified actions.
At the same time, as with other web automation and agentic systems, there is a risk that similar techniques could be misused for large-scale automated interactions or unintended manipulation if deployed irresponsibly.
These concerns are not unique to our method and are common to web-based agents in general.
We believe that appropriate safeguards, usage policies, and human oversight are essential for real-world deployment.
Overall, we do not foresee any novel ethical concerns introduced by this work beyond those already present in existing web agent and automation research.

\nocite{langley00}

\bibliography{references}
\bibliographystyle{icml2026}

\newpage
\appendix
\onecolumn
\section{Action Space}

\begin{table}[h]
    \centering
    \caption{Action space.}
    \label{tab:action_space}
    \small
    \setlength{\tabcolsep}{6pt}
    \renewcommand{\arraystretch}{1.2}
    \begin{tabular}{l l p{7.5cm}}
        \toprule
        \textbf{Action Type} & \textbf{Parameters} & \textbf{Description} \\
        \midrule
        click & [id] & Click on the webpage element specified by the given element id. \\
        \hline
        type & [id], [content] & Type the provided content into the input field with the specified id. By default, the Enter key is pressed after typing unless disabled. \\
        \hline
        hover & [id] & Hover the mouse cursor over the webpage element specified by the given id. \\
        \hline
        press & [key\_comb] & Simulate pressing a keyboard key or key combination (\eg, Ctrl+V). \\
        \hline
        scroll & [up/down] & Scroll the webpage vertically in the specified direction. \\
        \hline
        new\_tab & -- & Open a new, empty browser tab. \\
        \hline
        tab\_focus & [tab\_index] & Switch focus to the browser tab specified by its index. \\
        \hline
        close\_tab & -- & Close the currently active browser tab. \\
        \hline
        goto & [url] & Navigate the browser to the specified URL. \\
        \hline
        go\_back & -- & Navigate to the previously visited webpage. \\
        \hline
        go\_forward & -- & Navigate to the next webpage if a previous \texttt{go\_back} action was performed. \\
        \hline
        stop & [answer] & Terminate the task execution and return the final answer when the task objective is completed. \\
        \bottomrule
    \end{tabular}
\end{table}

We summarize the complete action space available to the web agent in Table~\ref{tab:action_space}, which is shared across all compared methods.

\section{Implementation Details}

\paragraph{Benchmarks and Baselines}

Our implementation is built upon the VisualWebArena \cite{koh2024visualwebarena} codebase, which provides the infrastructure for agent–webpage interaction.
Based on this framework, we implement our agent to execute tasks in both the VisualWebArena and Online-Mind2Web \cite{deng2023mind2web, xue2025an} benchmarks.
For baselines, ReAct \cite{Yao2023} is implemented using the official implementation provided in the VisualWebArena codebase.
WebDreamer \cite{Gu2025WebDreamer} is reproduced based on the code released by the original authors, and we strictly follow the configurations and design choices described in the original paper to ensure a fair comparison.

\paragraph{Foundation Models}

We use Qwen3-VL-Plus \cite{Qwen3-VL,Qwen2.5-VL,Qwen2-VL,Qwen-VL} as the foundation vision–language model for all agents.
Since Qwen3-VL-Plus is continuously updated, we specify a time-based snapshot for reproducibility.
All experiments in this paper are conducted using the model snapshot released on December 19, 2025.
For components in the benchmark evaluation pipeline that require model-based assistance, we adopt different models depending on modality.
Specifically, we use Qwen3-Max \cite{qwen3,qwen2.5,qwen2} for text-only evaluation tasks, and Qwen3-VL-Plus for evaluation steps involving visual inputs.

\paragraph{Hyperparameters}

To control computational cost, we set the number of judge model scoring samples to 3 and perform a single rollout per candidate action for world model simulation.
All other hyperparameters follow the official settings of VisualWebArena and WebDreamer.
For components specific to \ourmethod, we set the maximum number of feedback-driven action refinement iterations to 3.
The confidence threshold is set to $\theta = 0.4$, and at each refinement round we retain the top-$3$ candidate actions ranked by the judge model’s confidence scores.


\end{document}